\documentclass[]{bmvc2k}
\usepackage{eccvabbrv}
\usepackage{colortbl}
\usepackage{orcidlink}
\usepackage{graphicx}
\usepackage{booktabs}
\usepackage{bm}

\usepackage[accsupp]{axessibility}  %

\usepackage{enumitem}
\usepackage{amssymb}
\usepackage{algpseudocode}
\definecolor{commgreen}{HTML}{009933}
\usepackage[linesnumbered,ruled,vlined]{algorithm2e}

\SetCommentSty{licommfont}

\setenumerate[1]{itemsep=0pt,partopsep=0pt,parsep=\parskip,topsep=2.5pt}
\setitemize[1]{itemsep=1pt,partopsep=0pt,parsep=\parskip,topsep=1.5pt}
\setdescription{itemsep=0pt,partopsep=0pt,parsep=\parskip,topsep=2.5pt}
\usepackage{booktabs}
\usepackage{multirow}
\usepackage{dsfont}
\usepackage{threeparttable}
\usepackage{amsmath}
\usepackage{amssymb}
\usepackage{microtype}
\usepackage{mathtools}
\usepackage{pifont}
\usepackage[geometry]{ifsym}

\usepackage[capitalize]{cleveref}
\Crefname{section}{Section}{Sections}
\crefname{section}{Sec.}{Secs.}
\Crefname{table}{Table}{Tables}
\crefname{table}{Tab.}{Tabs.}

\newcommand\bdtitle[1]{\noindent\textbf{#1}}
\newcommand\topr{\toprule[1.2pt]}

\newcommand\midr{\toprule[0.65pt]}

\renewcommand{\arraystretch}{1.05}

\usepackage{adjustbox}
\newcommand{\graynum}[1]{\textcolor{figgray}{\ding{#1}}}

\definecolor{tabgreen}{RGB}{59,125,35}
\definecolor{figgray}{RGB}{127,127,127}
\definecolor{tabred}{RGB}{192,0,0}
\definecolor{cvprblue}{rgb}{0.21,0.49,0.74}
\definecolor{bondiblue}{rgb}{0.0, 0.18, 0.65}
\definecolor{dodgerblue}{rgb}{0.12, 0.56, 1.0}
\definecolor{jade}{HTML}{0EB83A}
\definecolor{mygray}{gray}{.92}

\newcommand{\ssymbol}[1]{^{\@fnsymbol{#1}}}

\usepackage{rotating}

\begin{document}

\newcommand{\validset}{{\textit{validation} set}}
\newcommand{\trainset}{{\textit{training} set}}
\newcommand{\testset}{{\textit{test} set}}
\newcommand{\pddshort}{{TraIL}}
\newcommand{\rpddfull}{{Intra-Ring {\pddfull}}}
\newcommand{\cpddfull}{{Intra-Class {\pddfull}}}
\newcommand{\rpdd}{{R-{\pddshort}}}
\newcommand{\cpdd}{{C-{\pddshort}}}
\newcommand{\ourmodel}{TraIL-Det}
\newcommand{\pddfull}{Transformation-Invariant Local Feature}

\title{{\pddshort}-Det: {\pddfull} Networks for 3D LiDAR Object Detection with Unsupervised Pre-Training}

\addauthor{Li Li}{li.li4@durham.ac.uk}{1}{0000-0002-9392-7862}
\addauthor{Tanqiu Qiao}{tanqiu.qiao@durham.ac.uk}{1}{0000-0002-6548-0514}
\addauthor{Hubert P. H. Shum}{hubert.shum@durham.ac.uk}{1}{0000-0001-5651-6039}
\addauthor{Toby P. Breckon}{toby.breckon@durham.ac.uk}{1,2}{0000-0003-1666-7590}

\addinstitution{
 Department of Computer Science\\
 Durham University\\
 Durham, UK
\vspace{0.7cm}
}

\addinstitution{
 Department of Engineering\\
 Durham University\\
 Durham, UK
}

\runninghead{Li, Qiao, Shum, Breckon}{{\pddshort}-Det Nets for 3D LiDAR Object Detection}

\maketitle

\begin{abstract} 
  3D point clouds are essential for perceiving outdoor scenes, especially within the realm of autonomous driving. Recent advances in 3D LiDAR Object Detection focus primarily on the spatial positioning and distribution of points to ensure accurate detection. However, despite their robust performance in variable conditions, these methods are hindered by their sole reliance on coordinates and point intensity, resulting in inadequate isometric invariance and suboptimal detection outcomes.
  To tackle this challenge, our work introduces \textbf{\underline{Tra}}nsformation-\textbf{\underline{I}}nvariant \textbf{\underline{L}}ocal ({\pddshort}) features and the associated {\ourmodel} architecture. Our {\pddshort} features exhibit rigid transformation invariance and effectively adapt to variations in point density, with a design focus on capturing the localized geometry of neighboring structures. They utilize the inherent isotropic radiation of LiDAR to enhance local representation, improve computational efficiency, and boost detection performance. To effectively process the geometric relations among points within each proposal, we propose a Multi-head self-Attention Encoder (MAE) with asymmetric geometric features to encode high-dimensional {\pddshort} features into manageable representations. 
  Our method outperforms contemporary self-supervised 3D object detection approaches in terms of mAP on KITTI ($\mathbf{67.8}$, 20\% label, moderate) and Waymo ($\mathbf{68.9}$, 20\% label, moderate) datasets under various label ratios (20\%, 50\%, and 100\%). 
\end{abstract}

\section{Introduction}

LiDAR-based point clouds, comprising 3D positions and LiDAR intensity/reflectivity~\cite{Geiger2012,li2021durlara,sun2020scalability,nuscenes2019,li2023less}, are essential for interpreting outdoor environments, particularly in the context of autonomous vehicle perception systems. The realm of 3D object detection has seen significant progress, with a variety of strategies aimed at predicting 3D bounding boxes. Recent approaches have employed color information~\cite{wu2023virtual,chen2017multiview,liang2018deep}, range imagery~\cite{meyer2019lasernet, meyer2019sensora}, and Birds Eye View (BEV) projections~\cite{chen2017multiview,liang2018deep} to devise multi-modal techniques that merge inputs from LiDAR and other sensors to improve feature representation and detection accuracy.

Despite these advancements, these methods commonly face challenges with isometric invariance due to their reliance primarily on coordinates and intensity data, often leading to suboptimal detection results~\cite{zhou2018voxelnet,yang2021st3d,wu2023virtual}. These limitations are predominantly caused by inadequate translational invariance and issues such as occlusions or sparse observations at longer ranges~\cite{li2023memoryseg}, which adversely affect the spatial distribution of the data.

In this work, we aim to identify features that (1) capture the localized geometric structure of neighboring points, (2) are invariant to rotation and translation, and (3) function effectively in noisy LiDAR outdoor scenes. Although various methods meet some of these criteria individually~\cite{jiang2018pointsift, liang2012geometric, melia2023rotationinvariant}, they often do not satisfy all simultaneously. Given the necessity for higher-level features that can encapsulate local geometry and potentially include LiDAR-specific attributes such as intensity and reflectivity, we focus on Pointwise Distance Distribution (PDD) features~\cite{widdowson2022resolving,widdowson2023recognizing}. PDD features are noted for their exceptional ability to provide robust and detailed geometric representations of point clouds, effectively maintaining both rotational and translational invariance while capturing intricate details of local geometry.

However, the direct use of PDD features is impractical due to their high dimensionality and the substantial memory and storage demands they impose on large-scale point clouds~\cite{li2024rapidseg}. Additionally, PDD tends to overlook local features because the inclusion of distant points can diminish the emphasis on nearby neighborhoods.

To facilitate 3D object detection, we propose the \textbf{\underline{Tra}}nsformation-\textbf{\underline{I}}nvariant \textbf{\underline{L}}ocal ({\pddshort}) features and the associated {\ourmodel} network. It leverages the robustness of {\pddshort} features against rigid transformations and variability in point cloud density, focusing on extracting compact features within defined local neighborhoods. Our approach employs inherent LiDAR isotropic radiation and multi-head self-attention to improve the representation of local features while reducing computational overhead. To effectively handle the high dimensionality of {\pddshort} features, we introduce a novel embedding method within our {\pddshort} Proposal Multi-head self-Attention Encoding ({\pddshort} MAE) module. Additionally, we enhance ability of the model to precisely localize individual objects and accurately identify different object categories through a joint optimization of discrimination and separation. This integration into the overall network is designed to elevate performance and expand generalization capabilities.

We conduct extensive experiments on KITTI~\cite{Geiger2012} and Waymo~\cite{sun2020scalability} datasets, where our methodology outperforms existing state-of-the-art (SoTA) self-supervised methods in 3D object detection.
Overall, our contributions are summarized as follows:
\begin{itemize}
\item A novel \textbf{Transformation-Invariant Local ({\pddshort})} feature for 3D object detection that ensures robustness to rigid transformations through isometry-invariant metrics.
\item A novel method for \textbf{embedding {\pddshort} with Multi-head self-Attention Encoder (MAE)} to capture the geometric relations between points, jointly attending to information from different representation subspaces at different positions.
\item A novel \textbf{open-source pre-training architecture {\ourmodel}}~\footnote{The code is publicly available at: \url{https://github.com/l1997i/rapid_seg}.} and supporting training methodology for 3D object detection that outperform the recent contemporary approaches of ProposalContrast~\cite{yin2022proposalcontrast}, DepthContrast~\cite{zhang2021self}, and PointContrast~\cite{xie2020pointcontrast}.

\end{itemize}
\section{Related Work}

\bdtitle{3D LiDAR Object Detection}: 
Initial approaches in the field convert LiDAR point clouds into 2D formats, specifically Bird Eye View (BEV) or range-view images~\cite{chen2017multiview,beltran2018birdnet} to facilitate 3D object detection. More recent advancements have shifted towards using voxel-based sparse convolution techniques~\cite{yan2018second,he2020structure,deng2021voxel,lang2019pointpillars} and point-based methodologies for set abstraction~\cite{shi2020pv,shi2019pointrcnn,yang20203dssd,yang2019std} to create more effective detection frameworks. A common challenge with LiDAR data is its low resolution for objects at a distance, leading to sparse detection outcomes. To mitigate this, researchers have delved into multimodal 3D object detection, demonstrating that the integration of LiDAR with RGB image data improves detection performance. Initial strategies enrich LiDAR points with image data~\cite{sindagi2019mvx, vora2020pointpainting, wang2021pointaugmenting}, while others pursue independent encoding of multimodal features, followed by their fusion either within the local Region of Interest (RoI)~\cite{chen2023futr3d,ku2018joint} or on the BEV plane~\cite{liu2023bevfusion}. Recent advancements utilize virtual points for feature fusion~\cite{wu2022sparse,yin2021multimodal}, which effectively improve the geometric representation of distant objects through depth estimation, showcasing significant promise for elevated detection performance. Nonetheless, virtual points introduce challenges related to their density and noise levels. VirConv~\cite{wu2023virtual} integrates RGB image data through virtual points and introduces StVD and NRConv as effective solutions to address the related challenges.

\bdtitle{Pointwise Distance Distribution}: 
Pointwise Distance Distribution (PDD) quantifies the local context of each point within a unit cell by sequentially measuring distances to nearby points. This isometry-invariant technique, developed by Widdowson \& Kurlin~\cite{widdowson2022resolving}, effectively addresses data ambiguity in periodic crystals, as demonstrated through detailed pairwise comparisons of atomic 3D clouds from structured periodic environments~\cite{widdowson2022resolving,widdowson2022average,widdowson2023recognizing}. Despite its proven effectiveness in periodic crystals and atomic clouds, PDD has not yet been applied to outdoor 3D point clouds. In outdoor scenarios, commonly used invariant features~\cite{liang2012geometric,jiang2018pointsift,melia2023rotationinvariant,li2022rotationinvariant} often struggle with irregular and sparse data, compounded by increased noise and environmental complexity~\cite{liang2012geometric,li2022rotationinvariant}. Additionally, the computational intensity of these features limits their feasibility for large-scale outdoor applications~\cite{jiang2018pointsift,melia2023rotationinvariant}. For instance, Melia~\etal~\cite{melia2023rotationinvariant} report a rotation-invariant feature that has difficulty scaling across diverse point cloud densities and sizes due to its computational demand and vulnerability to outdoor interferences. Recognizing these limitations, we propose the exploration of PDD features in outdoor settings, where accurately representing the local context of points in a transformation-invariant and structurally sound manner is paramount. Leveraging the architectural advantages of PDD, we introduce the {\pddshort} feature, specifically designed for LiDAR-based point clouds to adeptly capture the local geometric configuration of neighboring structures.

\bdtitle{Self-Supervised Learning Methods for Point Cloud}: 
Self-supervised Learning (SSL)~\cite{hjelm2018learning,chen2020simple,he2020momentum,wang2021exploring,yin2022semi} have demonstrated exceptional performance on various tasks, at times outperforming supervised methods. This work presents a proposal-level pretraining approach specifically designed for point cloud object detection. Simultaneously, studies like PointContrast~\cite{xie2020pointcontrast}, DepthContrast~\cite{zhang2021self}, ProposalContrast~\cite{yin2022proposalcontrast}, GCC-3D~\cite{liang2021exploring}, and STRL~\cite{huang2021spatio} explore the utility of contrastive SSL in point cloud pretraining. However, these methods encounter several issues. Firstly, some~\cite{zhang2021self,huang2021spatio} treat the entire point cloud scene as a single instance, overlooking the multiple object instances typically present~\cite{chen2021multisiam, zhao2021self}. Secondly, methods like~\cite{xie2020pointcontrast,liang2021exploring} focus on point-/voxel-level discrimination, which hinders the development of object-level representations crucial for 3D object detection. Thirdly, several approaches~\cite{xie2020pointcontrast,zhang2021self,huang2021spatio} ignore the semantic relationships between instances, concentrating instead on low-level details rather than more informative high-level patterns. While an additional self-clustering strategy is implemented to capture semantic features~\cite{liang2021exploring}, it supervises only moving voxels, which are too sparse to encompass all potential object candidates, resulting in a cumbersome two-stage training pipeline for separate 3D and 2D encoders~\cite{liang2021exploring}. %

\section{{\pddshort}-Det for 3D Object Detection Pre-Training}
\begin{figure}[tp]
    \centering
    \includegraphics[width=1\linewidth]{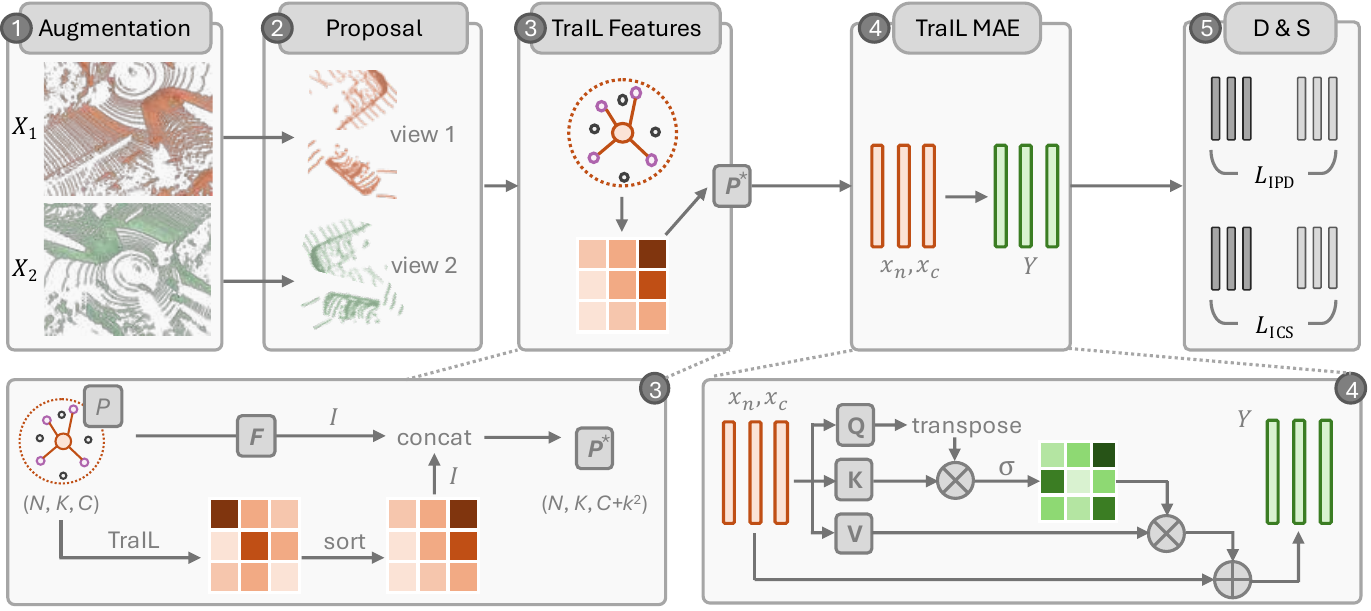}
    \caption{Our proposed \textbf{{\pddshort} architecture for 3D object detection} leverages {\pddshort} features from the point cloud. \graynum{202} We take point cloud inputs as input and augment them with differing views. \graynum{203} The augmented point clouds are sampled to the initial paired region proposals. \graynum{204} The encoding module ({\pddshort} MAE) extracts expressive proposal representations by considering the geometric relations among points within each proposal. \graynum{205} We extract the concatenated features with the Multi-Head Attention Encoding Module ({\pddshort} MAE). \graynum{206} Inter-Proposal Discrimination (IPD) and Inter-Cluster Separation (ICS), \ie D\&S module~\cite{yin2022proposalcontrast} are subsequently enforced to optimize the whole network.}
    \label{fig:overview}
\end{figure}

As shown in~\cref{fig:overview}, we propose {\pddshort}-Det architecture for 3D object detection which leverages {\pddshort} features from the 3D point cloud. The proposed architecture generates the {\pddshort}-based proposals from different augmented views (\cref{sec:trail_overview}). The {\pddshort}-based proposals are further embedded from {\pddshort} MAE (\cref{sec:trail_mae}) to process the geometric relations among points within each proposal.

\subsection{{\pddfull} ({\pddshort}) Overview}
\label{sec:trail_overview}

As illustrated in~\cref{fig:overview}~\graynum{202}\graynum{203}\graynum{204}, we present an overview of our {\pddshort} features in terms of their geometric descriptor (\cref{sec:descriptor}) and the augmented {\pddshort} proposal (\cref{sec:proposal}).

\subsubsection{{\pddshort} Descriptor}
\label{sec:descriptor}
Spatial transformations, including translation and rotation, are ubiquitous in real-world scenes, necessitating rotation-invariant representation of 3D point clouds. While translation-invariance can be achieved through weight sharing in 2D image understanding, 3D rotation-invariance remains a challenge due to the complexity of 3D geometry.

To achieve a 3D \textbf{\underline{Tra}}nsformation-\textbf{\underline{I}}nvariant \textbf{\underline{L}}ocal (TraIL) representation, we consider a point cloud patch $X \in \mathbb{R}^{3 \times K}$ with $K$ points, where each point $x_i$ ($i = 1, \ldots, K$) represents the 3D coordinates in Euclidean space. We define a transformation-invariant mapping $\text{TraIL}: \mathbb{R}^{3 \times K} \rightarrow \mathbb{R}^{C \times K}$, where $C \in \mathbb{N}^+$, to yield a consistent descriptor for geometrically identical point clouds under different orientations. The mapping should satisfy~\cref{equ:translation_invariant_condit}:
\begin{equation}
\label{equ:translation_invariant_condit}
\text{TraIL}(X) = \text{TraIL}(R X + T),
\end{equation}
where $\text{TraIL}(\cdot)$ denotes the translation-invariant operation, and $\text{TraIL}(X)$ is the invariant descriptor of $X$.

The difficulty in achieving 3D rotation-invariance has led us to seek alternative approaches that can effectively capture the intrinsic properties of 3D point clouds. Inspired by the concept of isometric invariance, which states that the properties of an object remain unchanged under rigid transformations, we hypothesize that features invariant to translation and rotation can be extracted from the geometric structure of point clouds. This idea is motivated by the fact that the distances between adjacent points in a point cloud remain constant regardless of the object orientation or position in 3D space. Therefore, we propose to exploit the Point Distance Distribution (PDD)~\cite{widdowson2022resolving,widdowson2023recognizing} as a translation-invariant feature, which measures the distribution of distances between adjacent points in a point cloud.

\textls[-2]{PDD is defined for a point cloud patch $X$ with $K$ points where $K > k$ and $k$ is the count of the nearest neighbours of a point, forming an $K \times k$ matrix $\text{PDD}(X; k)$. Each row $i$ of this matrix includes the ordered distances from the $i$-th point in $X$ to its $k$ nearest neighbours. Although the points in $X$ and rows of $\text{PDD}(X; k)$ are unordered, they are stored in lexicographic order to keep points and PDD matrix permutation-invariant (refer to the \textit{Supplementary Materials} for more details on computing PDD of point cloud and the corresponding \textit{sort} method).}

Considering $\text{PDD}$ properties under rotations $R$ and translations $T$, the distances between points in a point cloud remain unchanged. Subsequently, for a transformed cloud $X$ expressed as $RX + T$, the internal distances between points in $X$ and $RX + T$ are identical, preserving the Euclidean distance invariance under rotation and translation.
Consequently, the ordered distances from any point in $X$ to its $k$ nearest neighbours remain the same in both $X$ and $RX + T$, which leads to~\cref{equ:pdd_translation_invariant}:
\begin{equation}
\label{equ:pdd_translation_invariant}
\text{TraIL}(X)=\text{TraIL}(R X+T)=\text{PDD}(X ; k),
\end{equation}
\textls[-5]{where $\text{TraIL}(X)$ effectively captures the invariant spatial relationships within the cloud. Thus, $\text{PDD}(X; k)$ is a suitable candidate for $\text{TraIL}(X)$, fulfilling the requirements for a transformation-invariant 3D data representation regardless of point cloud orientation or position in space.}

\subsubsection{{\pddshort} Proposal}
\label{sec:proposal}
In 2D representation learning, some SSL methods utilize image proposals delineated by 2D bounding boxes. However, directly applying 3D bounding boxes to represent proposals in point clouds is impractical due to the vast candidate space and high computational cost of 3D spatial operations. Instead, we opt for spherical proposals.

In~\cref{fig:overview}, starting with the initial point cloud ${X}_0$, we remove road plane points to minimize background sampling~\cite{bogoslavskyi2017efficient}. We then apply farthest point sampling (FPS)~\cite{qi2017pointneta} to select $N$ distinct points from ${X}_0$, which serve as the centers for $N$ spherical proposals. Each proposal is formed by gathering $K$ nearby points within a predetermined radius $r$, ensuring proposal diversity. It results in two sets of spherical proposals ${P}_{1}$ and ${P}_{2}$, derived from two augmented views of ${X}_0$, represented as ${P}_{1}\in{X}_{1}$ and ${P}_{2}\in{X}_{2}$. We compute the {\pddshort} features $U_1$ and $U_2$ inside ${P}_{1}$ and ${P}_{2}$, \ie, $U_1 = \text{TraIL}(P_1)$, and $U_2 = \text{TraIL}(P_2)$. Since $P_1$ and $P_2$ may contain different numbers of points, the resulting {\pddshort} matrices $U_1$ and $U_2$ may also vary in size. Following the approach used for image matrices, we scale $U_1$ and $U_2$ to a predefined fixed size using bilinear interpolation $I(\cdot)$, making them suitable for input into subsequent neural networks.

\subsection{{\pddshort} Multi-Head Self-Attention Encoding ({\pddshort} MAE)}
\label{sec:trail_mae}

For the point cloud scene $X$ and corresponding proposals $P$, we first derive a global scene-wise representation using a backbone network, \eg, VoxelNet~\cite{zhou2018voxelnet} or PointNet++~\cite{qi2017pointneta}, denoted as $\boldsymbol{F}=f_{\text {Bbone }}(X)$.
Initial representations for the proposals $P^{*} \in \mathbb{R}^{N \times K \times C}$, are obtained by applying a bilinear interpolation function $I(\cdot)$ over $\boldsymbol{F}$, formulated as $P^{*}=I(P, \boldsymbol{F}) \oplus I(P, \boldsymbol{U})$, where $\oplus$ is the \texttt{concatenate} operator, $N$ is the number of proposals per view, $K$ is the number of points within a proposal, and $C$ is the channel number from the backbone network.

As shown in~\cref{fig:overview}~\graynum{205}, we employ the multi-head attention mechanism to process the geometric relations among points within each proposal. For each proposal $\boldsymbol{p} \in P^{*}$, with the size of $K \times C$, we designate the center point feature $\boldsymbol{x}_c \in \mathbb{R}^{1 \times C}$ of the proposal $\boldsymbol{p}$ as the query, recognizing its informativeness. Neighbor features $\boldsymbol{x}_n \in \mathbb{R}^{K \times C}$, derived from $\boldsymbol{p}$, serve as keys, with their differences to $\boldsymbol{x}_c$ encoding the asymmetric geometric relations.
Mathematically, the $\boldsymbol{x}_c$ and $\boldsymbol{x}_n$ are projected to query $\mathbf{Q}$, key $\mathbf{K}$, and value $\mathbf{V}$ embeddings:
\begin{equation}
    \mathbf{Q}=\delta\left(\boldsymbol{x}_c\right), \quad \mathbf{K}=\theta\left(\boldsymbol{x}_n-\boldsymbol{x}_c\right), \quad \mathbf{V}=\gamma\left(\boldsymbol{x}_n-\boldsymbol{x}_c\right),
\end{equation}
where $\delta, \theta$, and $\gamma$ represent the linear transformations.

The embeddings $\mathbf{Q}$, $\mathbf{K}$, and $\mathbf{V}$ are then processed by multi-head self-attention mechanism. In a $H$-head attention situation, $\mathbf{Q}$, $\mathbf{K}$, and $\mathbf{V}$ are further divided into $\mathbf{Q} = [\mathbf{Q}_1, \cdots , \mathbf{Q}_H]$, $\mathbf{K} = [\mathbf{K}_1, \cdots , \mathbf{K}_H]$, and $\mathbf{V} = [\mathbf{V}_1, \cdots , \mathbf{V}_H]$. 
For each $h$ ranging from $1$ to $H$, $\mathbf{Q}_h, \mathbf{K}_h, \mathbf{V}_h \in \mathbb{R}^{N \times D'}$ with $D' = D/H$. The output of the multi-head self-attention is computed as follows:
\begin{equation}
S^{(\mathrm{att})}\left(\mathbf{Q}, \mathbf{K}, \mathbf{V}\right)=\operatorname{softmax}\left(\frac{\mathbf{Q}_h\ \mathbf{K}_h^{\top}}{\sqrt{D'}}\right) \cdot \mathbf{V}.
\end{equation}

\begin{figure}[tb]
    \centering
    \begin{minipage}{0.6\linewidth}
        \centering
        \includegraphics[width=\linewidth]{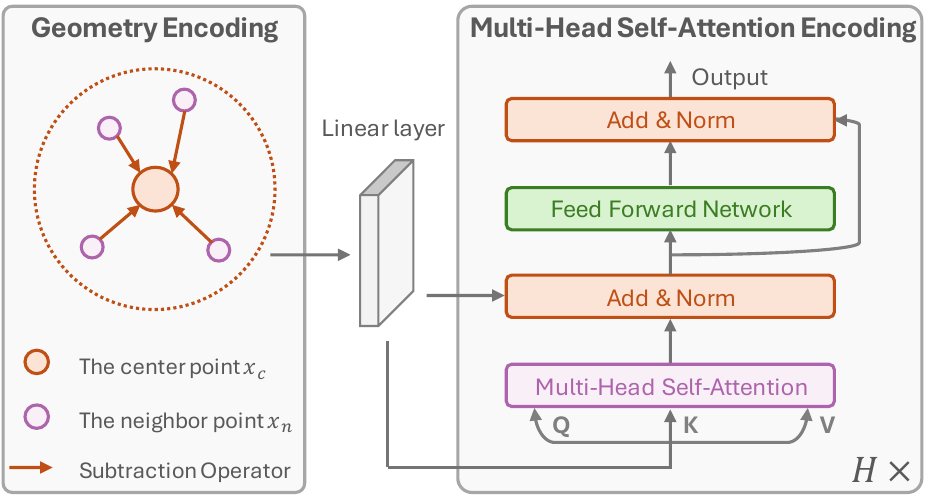}
    \end{minipage}\hfill
    \begin{minipage}{0.35\linewidth}
        \caption{\textls[-30]{\textbf{Multi-attention geometric encoding}. The asymmetric geometric features are computed from the proposal $P^{*}$, specifically the center and neighbor points, through a subtraction operator. The geometric features are further refined by a proposal-aware encoding module that utilizes a multi-head self-attention mechanism.}}
        \label{fig:self_atten_encode}
    \end{minipage}
\end{figure}

As shown in~\cref{fig:self_atten_encode}, a simple Feed-Forward Network (FFN) and residual operator are then adopted to obtain proposal representations as:
\begin{equation}
\boldsymbol{Y} = S^{(\mathrm{emb})}(\boldsymbol{x}_c, \boldsymbol{x}_n)=\mathcal{Z}\left(\mathcal{N}\left(\mathcal{Z}\left(\mathbf{S}^{(\text {att })}(\mathbf{Q}, \mathbf{K}, \mathbf{V})\right)\right)\right)
\end{equation}
where $\mathcal{Z}(\cdot)$ denotes add and normalization operator, $\mathcal{N}(\cdot)$ denotes a FFN with $2\times$ linear layers and $1\times$ ReLU activation. We observe that a stack of 3 identical self-attention
encoding modules (\ie, $H=3$) is ideal for our {\pddshort}-Det framework.

By implementing these operations across each region proposal in $X_1$ and $X_2$, we obtain refined proposal representations, $\boldsymbol{Y}_1, \boldsymbol{Y}_2 \in \mathbb{R}^{N \times C}$, which are further optimized through joint training for both inter-proposal discrimination and inter-cluster separation.
This integration of multi-head attention not only enriches the geometric understanding of the proposals but also enhances the model capacity to represent complex spatial dependencies effectively.

We follow ProposalContrast~\cite{yin2022proposalcontrast}, taking $\boldsymbol{Y_1}$ and $\boldsymbol{Y}_2$ to the D\&S module (\cref{fig:overview}~\graynum{206}), to optimize with inter-proposal discrimination (IPD) and inter-cluster separation (ICS) simultaneously in a self-supervised manner for better overall discriminative and classificatory capabilities. The hyper-parameter settings (\eg, $\alpha$, $\beta$, and $\tau$) for the D\&S module are consistent with those used in ProposalContrast~\cite{yin2022proposalcontrast}.

\section{Evaluation}
\label{sec:experiments}
\textls[-3]{We follow the standard SSL experimental framework, which involves pretraining a backbone network on extensive unlabeled data and subsequently fine-tuning this pretrained model on downstream tasks using a smaller set of labeled data. 
Unlike some previous 3D SSL approaches that utilize the ShapeNet~\cite{li2021shapenet} and ScanNet~\cite{dai2017scannet} datasets for pretraining—thereby concentrating exclusively on indoor environments and encountering significant domain gaps when applied to self-driving scenarios—we employ a different strategy to mitigate this limitation.
}

\subsection{Experimental Setup}

\bdtitle{Datasets}: We evaluate the transferability of our pre-trained model by pre-training on Waymo Open Dataset (WOD)~\cite{sun2020scalability} then fine-tuning on KITTI dataset~\cite{Geiger2012}. WOD comprises 798 training scenes (158,361 frames) and 202 validation scenes (40,077 frames), which is about $20\times$ larger than KITTI. We leverage the entire WOD training set for pretraining various 3D backbone architectures, explicitly avoiding the use of labels. KITTI contains 7,481 labeled samples, which are divided into two groups, \ie a training set (3,712 samples) and a validation set (3,769 samples). 

\bdtitle{Evaluation Protocol}: Mean Average Precision (mAP) and Mean Average Precision weighted by Heading (mAPH) with 40 recall positions ($R_{40}$) are employed to evaluate detection performance. We report results on the two difficulty levels and 3 classes, with 3D Intersection over Union (IoU) thresholds set at 0.7 for cars and 0.5 for pedestrians and cyclists.

\bdtitle{Implementation Details}: 
\textls[-14]{We consider four types of widely-used data augmentations to generate different views, \ie, random rotation ([$-\pi, +\pi$]), random scaling ($[0.5, 1.5]$), random flipping (X-axis, Y-axis), and point-wise random drop out. All experiments are conducted on $4\times$ NVIDIA A100 GPUs (1× for inference). Except for the parameters mentioned in~\cref{sec:ablation}, we follow the configurations from ProposalContrast~\cite{yin2022proposalcontrast} to facilitate comparison with SoTA approches.}

\subsection{Experimental Results of Transfer Learning}

\textls[-18]{We explore the effectiveness of self-supervised pre-training within the context of autonomous driving. We evaluate our methodology on multiple popular LiDAR point cloud datasets for autonomous driving, \ie, KITTI~\cite{Geiger2012} and WOD~\cite{sun2020scalability}. Our evaluation involves a comparative analysis of our {\pddshort}-Det against SoTA pre-training strategies~\cite{xie2020pointcontrast,zhang2021self,yin2022proposalcontrast} by fine-tuning detection models on these datasets. We utilize varying amounts of labeled data for fine-tuning to demonstrate the data efficiency of our approach. We employ a range of contemporary 3D object detectors as well to illustrate the broad applicability and generalizability of our pre-trained models.}
\begin{table}[thp]
\caption{{\textbf{Data-efficient 3D Object Detection on KITTI.} We pre-train the backbones of PointRCNN~\cite{shi2019pointrcnn} and PV-RCNN~\cite{shi2020pv} on Waymo and transfer to KITTI 3D object detection with different label configurations. Consistent improvements are obtained under each setting. Our approach outperforms all the concurrent self-supervised learning methods, \ie, DepthContrast~\cite{zhang2021self}, PointContrast~\cite{xie2020pointcontrast}, ProposalContrast~\cite{yin2022proposalcontrast}, GCC-3D~\cite{liang2021exploring}, and STRL~\cite{huang2021spatio}.}}
\label{tb:kitti}
\centering
\renewcommand\arraystretch{1.1}
\resizebox{1\textwidth}{!}
{
\begin{tabular}{@{}c||c|c||c|ccc|ccc|ccc@{}}
\topr
		\rowcolor{cvprblue!15}
\textbf{Fine-tuning with}  && \textbf{Pre-train.} &\textbf{mAP}       &  \multicolumn{3}{c|}{\textbf{Car}}          & \multicolumn{3}{c|}{\textbf{Pedestrian}   }         & \multicolumn{3}{c}{\textbf{Cyclist}}         \\
		\rowcolor{cvprblue!15}
{\textbf{various label ratios}} &\multirow{-2}{*}{\textbf{Detector}} &\textbf{Schedule}
&\textbf{(Mod.)}  & \textbf{Easy}             & \textbf{Mod.}           & \textbf{Hard}         & \textbf{Easy}             & \textbf{Mod.}           & \textbf{Hard} &\textbf{Easy}             & \textbf{Mod.}           & \textbf{Hard}        \\ \hline\hline
  &  & Scratch & 63.51 & 88.64	&75.23	&72.47       & 55.49 	& 48.90 &	42.23    &85.41 &	66.39 	&61.74  \\
  &  & Prop.Con.~\cite{yin2022proposalcontrast} & {66.20} & 88.52	&77.02	&72.56 & 58.66 	&51.90 	&44.98           &90.27 	&69.67 	&65.05     \\
\multirow{-1}{*}{\begin{minipage}[t]{3cm}\centering 20\% \\ ($\sim$ 0.7k frames)\end{minipage}}  &\multirow{-3}{*}{PointRCNN} & \ding{72} Ours& \textbf{67.80} & \textbf{89.07}	&\textbf{78.86}	& \textbf{73.63} & \textbf{59.12} 	&\textbf{53.37} 	& \textbf{46.11}           &\textbf{92.95} 	&\textbf{71.16} 	&\textbf{66.12}     \\ \cline{2-13}
&&  Scratch & 66.71 &91.81	&82.52	&80.11           &58.78 	&53.33 	&47.61   &86.74 	&64.28 	&59.53            \\
  &   & Prop.Con.~\cite{yin2022proposalcontrast}&{68.13} &\textbf{91.96}	&82.65	&80.15& \textbf{62.58} &	55.05 &	\textbf{50.06}           &\textbf{88.58} 	&66.68 	& \textbf{62.32}   \\ 
  & \multirow{-3}{*}{~~PV-RCNN} & \ding{72} Ours& \textbf{69.30} & 91.88 & \textbf{82.73}	& \textbf{80.39} & 62.22  & \textbf{56.94} & 49.85 & 88.43	& \textbf{68.24}  & 61.19 \\ \cline{2-12}   \hline

 &     &Scratch  & 66.73 & 89.12	&77.85	&75.36          &61.82 	&54.58 	& \textbf{47.90}   &86.30 	&67.76 	&63.26   \\
\multirow{4}{*}{\begin{minipage}[t]{3cm}\centering 50\% \\ ($\sim$ 1.8k frames)\end{minipage}}  & & Prop.Con.~\cite{yin2022proposalcontrast}&{69.23} &89.32	&79.97	&\textbf{77.39}  & 62.19 	&{54.47} 	&46.49          &\textbf{92.26} 	&73.25 	&68.51   \\
& \multirow{-3}{*}{PointRCNN} & \ding{72} Ours& \textbf{69.77} & \textbf{90.47}	& \textbf{81.23}  & 76.82  & \textbf{64.15}  &  \textbf{54.79}	& 47.28  & 91.16	& \textbf{73.29} &  \textbf{71.13} \\ \cline{2-13}
& & Scratch &69.63  & 91.77	 &82.68	&81.90           &63.70 	&57.10 	&52.77   &89.77 	&69.12 &	64.61           \\
  &  &Prop.Con.~\cite{yin2022proposalcontrast} &{71.76} &\textbf{92.29}	&82.92	&82.09& \textbf{65.82} 	&59.92 	&55.06        &91.87 	&72.45 	&\textbf{67.53}     \\
  & \multirow{-3}{*}{~~PV-RCNN} & \ding{72} Ours& \textbf{73.24} & 90.15	& \textbf{84.20}  & \textbf{85.01}  & 64.28  &  \textbf{61.43} & \textbf{56.09}  & \textbf{92.42} & \textbf{74.10} & 66.23 \\  \hline

   &    & Scratch & 69.45 & 90.02	&80.56	&\textbf{78.02} &62.59 	&55.66 	&48.69   &89.87 	&72.12 	&67.52   \\
   &    & DepthCon.~\cite{zhang2021self} & 70.26 &89.38	&80.32	&77.92          &65.55 	&57.62 	&50.98  &90.52 	&72.84 	&68.22   \\
\multirow{5}{*}{\begin{minipage}[t]{3cm}\centering 100\% \\ ($\sim$ 3.7k frames)\end{minipage}}  &  & Prop.Con.~\cite{yin2022proposalcontrast}&{70.71} &89.51	&80.23	&77.96  & 66.15 	&\textbf{58.82} 	&52.00            &91.28 	&73.08 	&68.45  \\
& \multirow{-4}{*}{PointRCNN} & \ding{72} Ours& \textbf{71.41} & \textbf{90.82}	& \textbf{81.95}  & 77.85  & \textbf{66.28}  &  58.73	& \textbf{53.96}  & \textbf{92.41} & \textbf{73.55}
  &  \textbf{71.53}  \\  \cline{2-13}
& &Scratch  & 70.57 & -	&84.50	&-          &- 	&57.06	&-   &- 	&70.14 	&-           \\
& &GCC-3D~\cite{liang2021exploring}  & 71.26 & -           & -        & -          &-   & -           &-  &-    &-           &-         \\
& &STRL~\cite{huang2021spatio}  & 71.46 & -          & 84.70        & -          &-    & 57.80            &-  &-    & 71.88            &-           \\
   &    & PointCon.~\cite{xie2020pointcontrast} & 71.55 &91.40	&84.18	&82.25          &65.73 	&57.74 	&52.46  &91.47 	&72.72 	&67.95   \\
  &  &Prop.Con.~\cite{yin2022proposalcontrast} &{72.92} & \textbf{92.45}	&84.72	&82.47  & \textbf{68.43} 	& 60.36 & \textbf{55.01} & 92.77 	&73.69 	&69.51      \\
  & \multirow{-6}{*}{~~PV-RCNN} & \ding{72} Ours& \textbf{73.89} & 92.10	& \textbf{85.39}  & \textbf{84.12}  & 68.01  &  \textbf{61.25} & 54.29  & \textbf{93.46} & \textbf{75.04}
  & \textbf{72.49}  \\ \hline

\end{tabular}
}
\vspace{-0.5cm}
\end{table}

\begin{figure}[tp]
    \centering
    \includegraphics[width=1\linewidth]{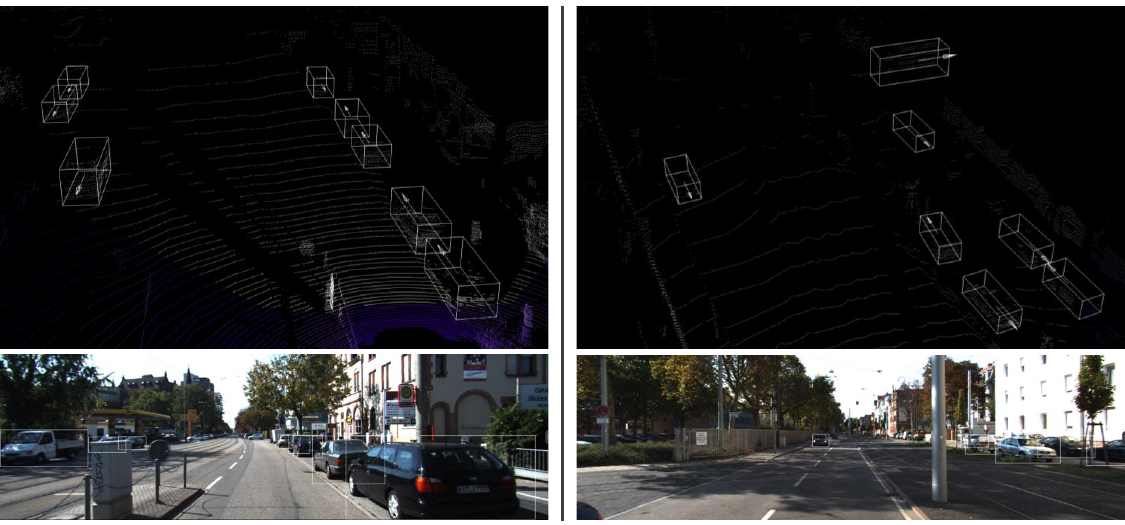}
    \caption{The qualitative results of 3D object detection with our {\pddshort}-Det on the KITTI dataset. The predicted 3D bounding boxes are marked within the point cloud frame, while the corresponding 2D bounding boxes are highlighted in the RGB images. In the point cloud visualization, white points represent those within the camera field of view (FOV), whereas purple points indicate those outside the camera FOV. Best viewed in color.}
    \label{fig:visual}
\end{figure} 

\subsubsection{Results on KITTI Dataset}
We evaluate the transferability of our pre-trained model by initially pre-training on Waymo and then fine-tuning on KITTI, using PointRCNN~\cite{shi2019pointrcnn} and PV-RCNN~\cite{shi2020pv} as baseline detectors. These detectors employ distinct 3D backbones (point-wise and voxel-wise), representing common types of 3D detectors. A key benefit of self-supervised pre-training is enhanced data efficiency, especially with limited annotated data -- we thus train the model with 20\% ($\sim 0.7$k), 50\% ($\sim 1.8$k) and 100\% ($\sim 3.7$k) labeled samples. \cref{tb:kitti} demonstrates that our pre-trained model boosts performance across both detectors compared to training from scratch and outperforms multiple existing methods. For instance, under the 50\% label setting, our model achieves 73.24\% mAP on moderate difficulty using PV-RCNN backbone, surpassing both the baseline and ProposalContrast~\cite{yin2022proposalcontrast} by 3.61\% and 1.48\% respectively. It also significantly outperforms DepthContrast~\cite{zhang2021self} and PointContrast~\cite{xie2020pointcontrast} in detecting cars (+1.40, average, 50\% moderate), pedestrians (+2.92, average, 50\% moderate) and cyclists (+3.30, average, 50\% moderate), highlighting its superior proposal-level representation and ability to handle imbalanced class distributions. Furthermore, we present supporting qualitative results in ~\cref{fig:visual} (more visualization results in the supplementary materials).

\subsubsection{Results on Waymo Open Dataset (WOD)}
We follow the widely-used OpenPCDet~\cite{openpcdet2020} protocol, fine-tuning the detectors on 20\% of the training data for 30 epochs. \cref{tb:waymo-D5} shows the results on Level-2 to other SoTA pre-training methods: GCC3D~\cite{liang2021exploring} and ProposalContrast~\cite{yin2022proposalcontrast}. Initially, we report results for training from scratch with different detectors, \ie, SECOND~\cite{yan2018second}, Part-A$^2$-Anchor~\cite{shi2020points}, PV-RCNN~\cite{shi2020pv}, and CenterPoint~\cite{yin2020center} (VoxelNet version), benchmarked against GCC3D~\cite{liang2021exploring}. Subsequently, we deploy our {\pddshort}-Det model with two widely-used detectors, \ie, CenterPoint~\cite{yin2020center} and PV-RCNN~\cite{shi2020pv} for evaluation.

\begin{table}[tp]
\caption{{\textbf{Comparisons between our model and other self-supervised learning methods on WOD}. All the detectors are trained by 20\% training samples following the OpenPCDet~\cite{openpcdet2020} configuration and evaluated on the validation set. Both PV-RCNN~\cite{shi2020pv} and CenterPoint~\cite{yin2020center} are used as baseline detectors.}}
\label{tb:waymo-D5}
\centering
\setlength\tabcolsep{6pt}
\renewcommand\arraystretch{1.1}
\resizebox{1\textwidth}{!}{
\begin{tabular}{@{}l||c||l|ccc@{}}
\topr
\rowcolor{cvprblue!15}  &\textbf{~Transfer} & \textbf{Overall}             & \textbf{Vehicle}           & \textbf{Pedestrian}            & \textbf{Cyclist}          \\  \rowcolor{cvprblue!15} \multirow{-2}{*}{\textbf{3D Object Detector}}
&\textbf{~Paradigm} & \textbf{AP/APH}             &  \textbf{AP/APH}           &  \textbf{AP/APH}            &  \textbf{AP/APH}          \\ \hline\hline
SECOND~\cite{yan2018second}                                    &Scratch &59.00/54.97            & 63.81/63.24	           & 56.77/46.66           & 56.42/55.02     \\
Part-A$^2$-Anchor~\cite{shi2020points}                                    &Scratch &64.81/61.63            & 69.04/68.49           & 58.21/50.56	          & 67.19/65.84
    \\
\hline

PV-RCNN~\cite{shi2020pv}                  &Scratch    & 60.88/57.20   & 66.12/65.50	   & 54.73/45.92	   & 61.77/60.20    \\
\ \ \textbf{+} GCC-3D~\cite{liang2021exploring}                  &Fine-tuning    & 61.30/58.18$_\text{{(+0.42/+0.98)}}$   & 65.65/65.10   & 55.54/48.02  & 62.72/61.43   \\
\ \ \textbf{+} Prop.Con.~\cite{yin2022proposalcontrast}    &Fine-tuning    & {62.62/59.28}$_\text{(+1.74/+2.08)}$ & 66.04/65.47   & 57.58/49.51   & 64.23/62.86    \\
\ \ \textbf{+} {\pddshort}-Det (\ding{72} Ours)                  &Fine-tuning    & {\textbf{64.16/60.62}}$_\textbf{{(+3.28/+3.42)}}$  & \textbf{67.26/66.73}   & \textbf{59.31/51.13}   & \textbf{65.90/64.01}    \\
\hline
CenterPoint~\cite{yin2020center}                  &Scratch    & 64.56/62.01   & 62.88/62.36   & 64.72/58.79 & 66.09/64.87    \\
\ \ \textbf{+} GCC-3D~\cite{liang2021exploring}                  &Fine-tuning     & 65.29/62.79$_\text{{(+0.73/+0.78)}}$  & 63.97/63.47   & 64.23/58.47   & 67.68/66.44    \\
\ \ \textbf{+} Prop.Con.~\cite{yin2022proposalcontrast}                 &Fine-tuning     & {66.42/63.85}$_\text{{(+1.86/+1.84)}}$   & 64.94/64.42    & \textbf{66.13/60.11}    & 68.19/67.01     \\
\ \ \textbf{+} {\pddshort}-Det (\ding{72} Ours)  &Fine-tuning    & {\textbf{66.55/ 63.94}}$_\textbf{{(+1.99/+1.93)}}$  &  \textbf{65.72/65.20}   &  65.49/59.53   &  \textbf{68.43/67.09}    \\
\hline
CenterPoint-Stage2~\cite{yin2020center}                  &Scratch    & 66.41/63.54 & 65.81/65.21   & 64.34/59.46   & 67.06/65.96   \\
\ \ \textbf{+} GCC-3D~\cite{liang2021exploring}                  &Fine-tuning     & 67.29/64.95$_\text{{(+0.88/+1.41)}}$  & 66.45/65.93   & 66.82/61.47   & 68.61/67.46    \\
\ \ \textbf{+} Prop.Con.~\cite{yin2022proposalcontrast}                 &Fine-tuning     & {68.06/65.69}$_\text{{(+1.65/+2.15)}}$   & 66.98/66.48    & 68.15/62.61   & \textbf{69.04/67.97}     \\
\ \ \textbf{+} {\pddshort}-Det (\ding{72} Ours)                    &Fine-tuning    & {\textbf{68.88/66.42}}$_\textbf{{(+2.47/+2.88)}}$  &  \textbf{68.21/67.67}   &  \textbf{69.50/63.76}   &   68.92/67.83   \\ \hline
\end{tabular}}
\end{table}

As demonstrated in~\cref{tb:waymo-D5}, our self-supervised pre-training significantly enhances the performance of popular 3D detectors. For PV-RCNN~\cite{shi2020pv}, our approach increases the APH by 3.42\% over training from scratch and surpasses SoTA ProposalContrast~\cite{yin2022proposalcontrast} by 1.34\% APH on average. Additionally, we apply our pre-training to CenterPoint~\cite{yin2020center} equipped with a VoxelNet backbone. The results indicate an improvement of 1.93\%. Moreover, utilizing our model with the two-stage CenterPoint architecture achieves an APH of 67.67\%, marking a 2.88\% increase over the model trained from scratch.

\subsection{Ablation Studies}  
\label{sec:ablation}

In~\cref{tab:ablation}, we ablate each component of our {\pddshort}-Det in depth. We pre-train the VoxelNet~\cite{zhou2018voxelnet} backbone on the full WOD~\cite{sun2020scalability} training set in an unsupervised manner, and evaluate the performance by finetuning the detector on WOD 20\% training data. We choose CenterPoint~\cite{yin2020center} which is trained from random initialization as the baseline.

\bdtitle{Effectiveness of {\pddshort} Feature}: In~\cref{tab:ablation}, we validate the efficacy of the proposed {\pddshort} feature. The use of the {\pddshort} feature result in a 60.52 mAP (+2.1). We further analysis how different neighbor sizes $k$ (\cref{equ:pdd_translation_invariant}) affected the {\pddshort}, finding $k=7$ optimal with a 62.29 mAP (+3.87). A larger $k$ can dilute local feature preservation by incorporating distant point distance, while a smaller $k$ risks losing important geometric relationships between neighboring points.

\bdtitle{Effectiveness of {\pddshort} MAE}: In~\cref{tab:ablation}, we evaluate the performance with (w/) and without (w/o) our proposed multi-head self-attention in our {\pddshort} MAE.
Employing the above mechanism, the results show a significant improvement of +1.77 mAP and +1.70 mAPH, which demonstrate the efficacy of the proposed {\pddshort} MAE with multi-head mechanism. 

\begin{table}[htp]
\caption{\textbf{Component-wise ablation} of our {\pddshort}-Det.}
\label{tab:ablation}
\centering
\setlength\tabcolsep{20pt}
\renewcommand\arraystretch{1.1}
\resizebox{\textwidth}{!}{
\begin{tabular}{c||c|c|cc}
\hline\midr
\rowcolor{cvprblue!15}
\textbf{Module}           & \textbf{Aspect}       & \textbf{Param.} & \textbf{mAP/mAPH} & $\Delta$ \\ \hline\hline
Baseline         & --           & --     & 58.42/55.64 & \\ \hline
\multirow{3}{*}{{\pddshort} Feature}&\multirow{2}{*}{Neighbor}& 5      &    59.48/57.19 &  (+1.06/+1.55)     \\
                 & \multirow{2}{*}{ size ($k$)}& 7      &    \textbf{62.29/59.89} &  \textbf{(+3.87/+4.25)}    \\
                 &              & 10     &    60.94/58.60  &  (+2.52/+2.96)   \\ \hline %
\multirow{2}{*}{{\pddshort} MAE}  & \multirow{1.25}{*}{Multi-head}   & w/     &   \textbf{62.29/59.89}  &   \textbf{(+3.87/+4.25)}   \\
                 & \multirow{1}{*}{attention} & w/o    &   60.52/58.19  &   (+2.10/+2.55)  \\ \bottomrule
\end{tabular}}
\end{table}    %
\section{Conclusion}
\label{sec:conclusion}

In conclusion, our \textbf{\underline{Tra}}nsformation-\textbf{\underline{I}}nvariant \textbf{\underline{L}}ocal (\pddshort) Features within the {\ourmodel} architecture effectively address the limitations of traditional 3D LiDAR object detection by focusing on localized geometry and relationships of points inside proposals. The introduction of the Multi-head self-Attention Encoder (MAE) efficiently processes and encodes high-dimensional {\pddshort} features, leveraging inherent LiDAR isotropic radiation for enhanced representation and computational efficiency. Experimentally, our approach outperforms existing self-supervised methods on KITTI~\cite{Geiger2012} and Waymo~\cite{sun2020scalability} datasets, demonstrating its effectiveness in advancing 3D object detection pre-training.

Our features are also highly effective for 3D semantic segmentation~\cite{li2024rapidseg}. Future research directions may include exploring more downstream tasks and applying the proposed features to domain adaptation. These features enhance robustness to transformations, making them well-suited for complex tasks such as domain adaptation. Additionally, their effectiveness in 3D semantic segmentation~\cite{li2024rapidseg} suggests they could improve performance across a broader range of downstream tasks.    %

\clearpage  %

\bibliography{abbr,ref}
\end{document}